\documentclass{article}

\usepackage[preprint]{nips_2018}
\usepackage[utf8]{inputenc} 
\usepackage[T1]{fontenc}    
\usepackage{hyperref}       
\usepackage{url}            
\usepackage{booktabs}       
\usepackage{amsfonts}       
\usepackage{nicefrac}       
\usepackage{microtype}      

\usepackage{graphicx}
\usepackage{mathabx}
\usepackage{caption}    
\newcommand{\KL}{{D_\mathrm{KL}}}
\newcommand{\xtest}{{x_\mathrm{test}}}
\renewcommand{\d}{{\,\mathrm{d}}}

\usepackage{booktabs,tabularx,caption}
\newcolumntype{C}{>{\centering\arraybackslash}X} 

\title{q-Space Novelty Detection with Variational Autoencoders}

\author{
Aleksei Vasilev\,$^{\text{1}}$,
Vladimir~Golkov\,$^{\text{1}}$,
Marc~Meissner\,$^{\text{1}}$,
Ilona Lipp\,$^{\text{2}}$,
Eleonora Sgarlata\,$^{\text{2,3}}$,\\
\textbf{
Valentina Tomassini\,$^{\text{2,4}}$,
Derek K. Jones\,$^{\text{2}}$,
Daniel~Cremers\,$^{\text{1}}$
}
\\
$^1$ Computer Vision Group, Technical University of Munich, Germany\\
$^2$ CUBRIC, Cardiff University, UK\\
$^3$ Department of Neurology and Psychiatry, Sapienza University of Rome, Italy\\
$^4$ Division of Psychological Medicine and Clinical Neurosciences, Cardiff University, UK\\
\footnotesize{\texttt{ \{alex.vasilev, vladimir.golkov, marc.meissner, cremers\}@tum.de, }}\\
\footnotesize{\texttt{ elesgarlata@hotmail.it, \{lippi, tomassiniv, jonesd27\}@cardiff.ac.uk}}
}

\begin{document}

\maketitle

\begin{abstract}
In machine learning, novelty detection is the task of identifying novel unseen data. During training, only samples from the normal class are available. Test samples are classified as normal or abnormal by assignment of a novelty score. Here we propose novelty detection methods based on training variational autoencoders (VAEs) on normal data. Since abnormal samples are not used during training, we define novelty metrics based on the (partially complementary) assumptions that the VAE is less capable of reconstructing abnormal samples well; 
that abnormal samples more strongly violate the VAE regularizer;
and that abnormal samples differ from normal samples not only in input-feature space, but also in the VAE latent space and VAE output.
These approaches, combined with various possibilities of using (e.g.~sampling) the probabilistic VAE to obtain scalar novelty scores, yield a large family of methods.
We apply these methods to magnetic resonance imaging, namely to the detection of diffusion-space (\mbox{q-space}) abnormalities in diffusion MRI scans of multiple sclerosis patients, i.e.~to detect multiple sclerosis lesions without using any lesion labels for training. Many of our methods outperform previously proposed q-space novelty detection methods. We also evaluate the proposed methods on the MNIST handwritten digits dataset and show that many of them are able to outperform the state of the art.

\end{abstract}

\section{Introduction} 

The purpose of novelty detection is to score how dissimilar each test sample is from a ``normal'' training set. This problem can also be seen as ``one-class classification'', where the model is trained to represent the normal class only. Application domains include medical diagnostic problems, fraud detection, and failure detection in industrial systems.

Novelty detection techniques can be classified into probabilistic, distance-based, reconstruction-based, domain-based, and information-theoretic~\cite{nd}. We here focus on the first three categories.

The quality of novelty detection results depends on the algorithm and data distribution. Deep generative neural networks can be applied to reveal internal structure of the data and learn a better data representation. In this paper we design a set of novelty detection methods based on variational autoencoders (VAEs). We apply them to directly detect abnormalities such as multiple sclerosis lesions in diffusion magnetic resonance imaging (diffusion MRI). Non-deep novelty detection methods for diffusion space (q-space) have been used to tackle this problem~\cite{qnd,qnd2018}, and achieved promising AUC scores between $0.82$ and $0.89$ in multiple sclerosis lesion segmentation on various datasets. We compare the proposed VAE-based novelty detection methods with the original q-space novelty detection, and show that some of them are able to perform better.

\subsection{Related work on novelty detection with generative models}
\label{rel_work}
With a dramatic increase of research in deep generative models during the last years, several new methods for novelty detection were proposed. These methods try to learn the normal patterns in the data using variational autoencoders, adversarial autoencoders~\cite{advae}, or generative adversarial networks (GANs)~\cite{gan}.

In~\cite{rprob} the VAE trained on a normal class only is used to detect abnormalities in the test data. The novelty score of the test sample is computed in the following way: the encoder of the VAE infers the distribution of the test sample in the latent space; several samples from this distribution are passed to the stochastic decoder of the VAE to infer a number of distributions of the test sample in the original feature space; these distributions are used to compute the average probability density value of the test sample, which is used as a novelty score.

A novelty score metric based on the adversarial autoencoder network is proposed in~\cite{aae}.
The model is trained to match the posterior distribution of the normal data in a latent space with a chosen prior distribution. The novelty metric in this case is based on the likelihood of the test sample according to the prior: low likelihood indicates that the test sample unlikely belongs to the normal data class.

Another approach~\cite{dagmm}, uses a framework consisting of an autoencoder network and a network that estimates a Gaussian mixture model of a normal class distribution in the latent space. For the test sample, both the reconstruction error of the autoencoder and the likelihood of this sample according to the estimated mixture model are used to define abnormality.

In~\cite{gandm, gandr}, a GAN is trained to learn the distribution of the normal data. During test time a search over the latent space is performed to find the closest generated sample to the test sample. Then this closest generated sample is fed into the discriminator and the overall novelty score is computed as a weighted sum of the discriminator loss and of the difference between the input and the generated sample.

Another line of work~\cite{safer} tries to unify classification and novelty detection into a single framework. The main concept is to train a separate class-conditional generative model (VAE or GAN) for each normal class in the training dataset. During test time, for each of these generative models a search over latent space is performed to find the distance between the test sample and the closest generated sample. The minimal distance across all class-conditional generators is a novelty score. If this score is below a certain threshold, then the test sample is classified with the class of the corresponding class-conditional generator. One of the methods we employ utilizes this idea for novelty detection, using \emph{only one} generative model for all the normal data, see Section~\ref{opt}. 

A GAN with a mixture generator is used for novelty detection in~\cite{gand}. In contrast to most novelty detection approaches, here the model is trained on both normal and abnormal data.

\subsection{Diffusion MRI}
\label{dmri}
Diffusion MRI is a magnetic resonance imaging (MRI) technique that uses the diffusion of water molecules to generate contrast in MR images. Since this diffusion is not free and affected by obstacles, it can reveal microstructural details about the tissue. For each of several chosen diffusion-space (q-space) coordinates, a diffusion-weighted 3D image is acquired.

\textbf{Classical data processing in diffusion MRI}\hspace{4pt}
Traditional diffusion MRI processing methods fit a handcrafted mathematical or physical model/representation to the measurements, and interpret the estimated model parameters. However, these approaches have several limitations, since the model-fitting procedure is ill-conditioned, and interpreting its results requires prior knowledge about how disease-related microstructural changes affect the model parameters. 

\textbf{Supervised and weakly-supervised deep learning in diffusion MRI}\hspace{4pt}
Recent research shows that deep learning can overcome said issues by learning a direct mapping between q-space measurements and diseases~\cite{tmi,qdl2018}.  In deep learning terminology, each diffusion-weighted 3D image corresponding to a certain q-space coordinate is treated as a \emph{``channel''} of the overall multi-channel 3D volume. For \emph{voxel-wise supervised learning}, for example to reconstruct missing q-space measurements from existing ones, or to predict handcrafted-model-based parameters more robustly and at a shorter scan time, or to directly estimate tissue types and properties, a ``voxels-to-voxels'' convolutional network can be used~\cite{tmi}. It can consist purely of convolutional layers with filter size $1\times 1\times 1$ if the risk of spatial bias (overfitting) should be excluded~\cite{tmi}. On the other hand, \emph{global supervised learning} (i.e.~image-wise rather than voxel-wise prediction) and \emph{voxel-wise weakly-supervised learning} can be performed with a convolutional network that reduces spatial resolution using pooling and/or fully-connected layers~\cite{qdl2018}. 
However, supervised and weakly-supervised disease detection requires disease-specific labels.

\textbf{Novelty detection in diffusion MRI}\hspace{4pt}
The aforementioned methods are complemented by q-space novelty detection methods, which do not require disease-related labels~\cite{qnd,qnd2018}. 
In this line of work, each voxel is treated as a separate $d$-dimensional feature vector, where $d$ is the number of measured diffusion directions. Voxels from scans of healthy volunteers are used as a reference dataset, and the Euclidean distance in feature space between the test datapoint and its nearest neighbor from the reference dataset is used as novelty score. A high novelty score thus indicates that the voxel is lesioned.
This novelty score coincides with multiple sclerosis lesions at AUC scores between $0.82$ and $0.89$ on various datasets, and we use this method as a baseline to compare our methods with.

\subsection{Our contributions}

In this paper we show that the usage of a variational autoencoder can help to better understand the normal patterns in the data and thus improve the quality of novelty detection. We further explore the possibilities of applying novelty detection to diffusion MRI processing. 
The main contributions of the paper can be summarized as follows:
\begin{itemize}
    \item We propose several new novelty detection methods in the VAE original and latent feature spaces. These methods can be applied to different novelty detection tasks. Our code is publicly available at \url{https://github.com/VAlex22/ND_VAE}
    \item We adapt the VAE network to q-space novelty detection. We show that this solution can beat the performance of the original q-space novelty detection algorithm.
\end{itemize}

\section{VAE-based Novelty Detection Methods}
\label{methods}

A~variatioal autoencoder is a deep neural network that models a relationship between a low-dimensional latent random variable $z$ and a random variable $x$ in the original data space. A~VAE consists of an encoder and a decoder. The encoder is trained to approximate the posterior distribution $q_\phi(z|x)$, where $\phi$ are the network parameters, learned with backpropagation during training. The decoder performs the inverse task: it approximates the posterior distribution $p_\theta(x|z)$, where $\theta$ are learned parameters. In contrast to the traditional autoencoder, the VAE has a stochastic encoder and decoder: their outputs for a given input are not the variables $z$ and $x$, but the parameters of the distributions of $z$ and $x$. 

The goal of the VAE is to learn a good approximation $q_\phi(z|x)$ of the intractable true posterior $p_\theta(z|x)$. 
The quality of the approximation can be evaluated with the Kullback--Leibler divergence: $\KL\big(q_\phi(z|x)\parallel p_\theta(z|x)\big)$. This divergence cannot be computed directly~\cite{vae1,vae2}, but it can be minimized by maximizing the sum of the Evidence Lower Bound (ELBO) on the marginal likelihood of datapoints~$x_i$: $\mathrm{ELBO}_i = E_{q_\phi(z|x_i)}[\log p_\theta(x_i|z)] - \KL\big(q_\phi(z|x_i)\parallel p(z)\big)$,
where $p(z)$ is a prior distribution of $z$ which is usually chosen as a unit Gaussian. The loss function of the VAE is then:
\begin{equation}\label{vae_loss}
\mathcal{L}_\mathrm{VAE} = -\sum_i\mathrm{ELBO}_i = -\sum_i \big[E_{q_\phi(z|x_i)}[\log p_\theta(x_i|z)] - \KL\big(q_\phi(z|x_i)\parallel p(z)\big)\big],\end{equation}
where the sum is calculated over all training data samples $x_i$.

From the machine learning perspective, the first term in Eq.~\eqref{vae_loss} can be considered as a reconstruction error forcing the model to reconstruct the input and the second term is a regularizer that prevents the model from giving each input a representation in a different region of the latent space.

In our case we train the VAE model to capture normal data only. Thereby, the VAE learns distributions of the normal dataset in latent and original feature space.  
Both of these distributions as well as their combination can be used to define novelty score metrics, thus we split our methods into three main categories. Novelty detection methods in latent space and original feature space are illustrated in Fig.~\ref{fig:pipeline}. 

\subsection{Novelty in the latent space}
The trained VAE maps each sample $x$ to a distribution {$z$} in some lower-dimensional latent space. This representation can be used to define several novelty metrics.
\subsubsection{Novelty as VAE regularizer}
\label{n_vae}
The VAE loss function includes the following regularizer term: $\KL\big(q_\phi(z|x) \parallel \mathcal{N}(0,I)\big)$. This term forces the model to map inputs closely to the unit Gaussian distribution $\mathcal{N}(0,I)$ in the latent space. Without this regularizer the model could learn to give each input a representation in a different region of the latent space. 

For the model trained on the normal class only, one could expect that abnormal samples will have distributions in the latent space that diverge more from the unit Gaussian than normal samples. We thus can use this term as a novelty score for test sample $\xtest$:
\begin{equation}
N_\textrm{VAE-reg}(\xtest) = \KL\big(q_\phi(z|\xtest)\ \parallel \mathcal{N} (0,I)\big).
\end{equation}

\subsubsection{Distance-based approaches in latent space}
\label{lat_dist}
The latent space of a VAE trained on the normal class can be considered an effective representation of the distribution of normal data. Therefore, classical novelty detection approaches can be adapted to be used in this space. The algorithm here is to construct the reference dataset by capturing normal data in the latent space. The latent representation of each test sample should also be inferred. Then we can use nearest neighbour analysis to find the closest sample from the reference dataset to the test sample in a latent space using some distance measure. The distance to this closest sample will be a novelty score. A~VAE maps each input point to a distribution (rather than a point) in latent space. We propose two distance measures:

\begin{figure}
  \centering
  \includegraphics[width=0.8\textwidth,trim={3cm 4.2cm 2.8cm 6.2cm},clip]{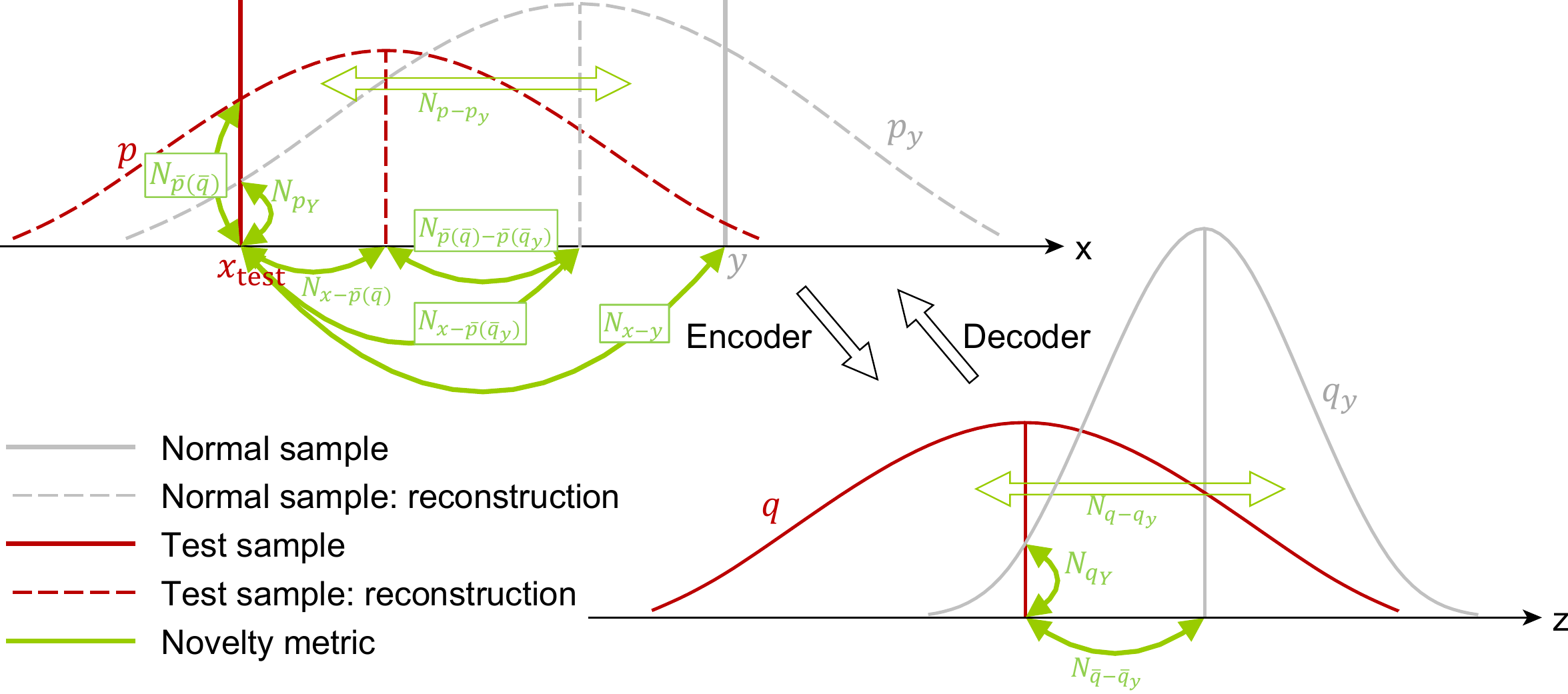}
  \caption{
  Proposed novelty detection methods for the simplified case of 1D original space $x$ and 1D latent space $z$ of the VAE with only one normal sample $y$ and one test sample $\xtest$. Arrows measure distances between points ($\curvearrowbotleftright$) or distributions ($\Leftrightarrow$). The scores $N_{x-\bar{p}(\bar{q})}$ and $N_{\bar{p}(\bar{q})}$ 
  quantify whether the VAE reconstructs $\xtest$ badly; $N_{x-y}$, $N_{x-\bar{p}(\bar{q}_y)}$ and $N_{\bar{p}(\bar{q})-\bar{p}(\bar{q_y})}$ measure how dissimilar $\xtest$ (or its reconstruction ``$\bar{p}(\bar{q})$'') is from normal samples $y$ (or their reconstruction ``$\bar{p}(\bar{q}_y)$''); $N_{q-q_y}$ and $N_{\bar{q}-\bar{q}_y}$ measure how dissimilar $\xtest$ is from $y$ in latent space; $N_{p_Y}$ and $N_{q_Y}$ measure how likely $\xtest$ belongs to the modeled distribution of normal data. In this illustration, methods that use stochastic estimates ``$\hat{q}\,$'' or ``$\hat{p}\,$'' correspond to the methods using deterministic means 
  ``$\bar{q}\,$'' or ``$\bar{p}\,$''. For example, $N_{x-\bar{p}(\bar{q}_y)}$, $N_{x-\bar{p}(\hat{q}_y)}$, $N_{x-\hat{p}(\bar{q}_y)}$, $N_{x-\hat{p}(\hat{q}_y)}$ all correspond to the same green arrow, but use distinct estimation methods.}
  \label{fig:pipeline}
\end{figure}

\begin{enumerate}
    \item \textbf{Euclidean distance between means of the distributions.} This approach uses only information about the means of the approximated posterior in the latent space for both normal and test datapoints. The novelty score in this case is computed as the distance between between the mean of the latent-space distribution of the test datapoint and the closest latent-space distribution mean of a normal sample:
    \begin{equation}
    N_{\bar{q}-\bar{q}_y}(\xtest) = \min_{y \in Y}\left\|E[q_\phi(z|\xtest)] - E[q_\phi(z|y)]\right\|_2^2,
    \end{equation}
    where the minimum is taken over all normal samples $y$ from the normal dataset $Y$.
    \item \textbf{Bhattacharyya distance between distributions.} The Bhattacharyya distance is a symmetric measure of dissimilarity of two probability distributions $p$ and $q$. It is defined as \(D_B(p,q) = -{\ln(\mathrm{BC}(p,q))}\), where \(\mathrm{BC}(p,q) = \int\sqrt{p(z)q(z)} \d z\) is the Bhattacharyya coefficient of distributions $p$ and $q$. This approach utilizes information about the full learned distributions, computing the amount of the overlap between them. The proposed novelty score is defined as the Bhattacharyya distance between the latent-space distribution $q_\phi(z|\xtest)$ of the test sample and the most similar latent-space distribution $q_\phi(z|y)$ of a normal sample:
    \begin{equation}N_{q-q_y}(\xtest) = \min_{y \in Y}D_B(q_\phi(z|\xtest), q_\phi(z|y)).\end{equation}
\end{enumerate}

\subsubsection{Density-based approach in latent space}
\label{lat_den}
Another approach to novelty detection is to estimate the density of normal data in the latent space. Each normal datapoint is represented as a Gaussian distribution in the VAE latent space. Thus the distribution of the whole normal dataset can be estimated as an average of these Gaussians: 
$q_Y(z)=\frac{1}{|Y|}\sum_{y \in Y} q_\phi(z|y)$. 
Then, the novelty score for the test sample can be computed from the density estimate $q_Y$ of the normal dataset, evaluated at the mean of the latent distribution $q_\phi(z|\xtest)$ of the test sample (see also Fig.~\ref{fig:pipeline}): 
\begin{equation}N_{q_Y}(\xtest) = -q_Y(E[q_\phi(z|\xtest)]).\end{equation}

\subsection{Novelty in the original feature space}

\subsubsection{VAE reconstruction-based approaches}
Like a traditional autoencoder network, a VAE is trained to reconstruct the input. If a VAE is trained on the normal class only, one could expect that it learns how to reconstruct data from the normal class well, but may not be particularly good at reconstructing unseen data, so the reconstruction error for the unseen class should be higher. Thus, the reconstruction error can be used as a novelty score.

The encoder and decoder are both stochastic. For each of them, we can either draw samples, or consider the mean. For the decoder it is also possible to consider the entire distribution and compute the density value of the test sample. We propose six possible reconstruction-based novelty scores:
\begin{enumerate}
    \item \textbf{Deterministic reconstruction error}: Use means of both the encoder \(q_\phi(z|x)\) and the decoder \(p_\theta(x|z)\) to deterministically reconstruct the input and compute the reconstruction error:
    \begin{equation}\label{dete}N_{x-\bar{p}(\bar{q})}(\xtest) = \left\|\xtest - E\big[p_\theta\big(x|E[q(z|\xtest)]\big)\big]\right\|_2^2.\end{equation} 
    \item \textbf{Deterministic reconstruction likelihood}: Compute log likelihood of the input given the mean of the encoder:
    \begin{equation}\label{detd}N_{\bar{p}(\bar{q})}(\xtest) = -{\log p_\theta\big(\xtest|E[q(z|\xtest)]\big)}.\end{equation} 
    \item \textbf{Encoder-stochastic reconstruction error}: Use samples from the encoder and mean of the decoder to compute several possible reconstructions of the input and calculate the average reconstruction error:
    \begin{equation}\label{esor}N_{x-\bar{p}(\hat{q})}(\xtest) = \underset{z_i \sim q_\phi(z|\xtest)}{\mathrm{mean}} \left\|\xtest - E[p_\theta(x|z_i)]\right\|_2^2.\end{equation}
    \item \textbf{Encoder-stochastic reconstruction likelihood}: Compute several possible log likelihood function values of the input given samples from the encoder:
    \begin{equation}\label{esod}N_{\bar{p}(\hat{q})}(\xtest) = \underset{z_i \sim q_\phi(z|\xtest)}{\mathrm{mean}} -{\log p_\theta(\xtest|z_i)}.\end{equation}
    \item \textbf{Decoder-stochastic reconstruction error}: Use mean of the encoder and samples from the decoder to compute average reconstruction error:
    \begin{equation}\label{dso}N_{x-\hat{p}(\bar{q})}(\xtest) = \underset{x_i \sim p_\theta(x|E[q_\phi(z|\xtest)])}{\mathrm{mean}} \left\|\xtest - x_i\right\|_2^2.\end{equation}
    \item \textbf{Fully-stochastic reconstruction error}: Use samples from both the encoder and the decoder and compute average reconstruction error:
    \begin{equation}\label{fs}N_{x-\hat{p}(\hat{q})}(\xtest) = \underset{\substack{x_i \sim p_\theta(x|z_i) \\ z_i \sim q_\phi(z|\xtest)}}{\mathrm{mean}} \left\|\xtest - x_i\right\|_2^2.\end{equation}
\end{enumerate}
The novelty score defined in Eq.~\eqref{esod} is equal to ``reconstruction probability'' proposed by~\cite{rprob}.

In metrics~\eqref{esor},~\eqref{esod},~\eqref{dso} and~\eqref{fs} the average operation can be replaced by the min operation, producing four more novelty metrics.

\subsubsection{Distance- and density-based approaches}
In addition to reconstruction-based approaches, it is possible to apply distance- and density-based approaches described in Sections~\ref{lat_dist} and~\ref{lat_den} to the distributions produced by the decoder of the VAE. Therefore we can get three more novelty metrics:
\begin{equation}\label{eud}N_{\bar{p}(\bar{q})-\bar{p}(\bar{q}_y)}(\xtest) = \min_{y \in Y} \Bigl\|
\underbrace{E\big[p_\theta\big(x|E[q_\phi(z|\xtest)]\big)\big]}_\text{Deterministic reconstruction of $\xtest$} -
\underbrace{E\big[p_\theta\big(x|E[q_\phi(z|y)]\big)\big]}_\text{Deterministic reconstruction of $y$}
\Bigr\|_2^2,\end{equation}
\begin{equation}\label{bhd}N_{p(\bar{q})-p(\bar{q}_y)}(\xtest) = \min_{y \in Y} D_B\Big(p_\theta\big(x|E[q_\phi(z|\xtest)]\big), p_\theta\big(x|E[q_\phi(z|y)]\big)\Big),\end{equation}
\begin{equation}\label{den}N_{p(\bar{q}_Y)}(\xtest) = - p_Y(E[p_\theta(x|E[q_\phi(z|\xtest)])]),\end{equation} 
where
$p_Y(x)=\frac{1}{|Y|}\sum_{y \in Y} p_\theta(x|E[q_\phi(z|y)])$
is the average of the distributions reconstructed from the normal-class samples in the original feature space.

Metrics in Eqs.~\eqref{eud} and~\eqref{den} can also be computed using the original test sample instead of its deterministic reconstruction produced by the VAE. Thus, two additional novelty scores are possible:
\begin{equation}\label{eudo}N_{x-\bar{p}(\bar{q}_y)}(\xtest) = \min_{y \in Y} \Bigl\|
\xtest - E\big[p_\theta\big(x|E[q_\phi(z|y)]\big)\big]\Bigr\|_2^2,\end{equation}
\begin{equation}\label{deno}N_{p(\bar{q}_Y)(x)}(\xtest) = - p_Y(\xtest).\end{equation} 
It is also possible to apply the Euclidean distance-based approach to the reconstructed test sample and original (not reconstructed) normal datapoints:
\begin{equation}\label{eudo2}N_{\bar{p}(\bar{q})-y}(\xtest) = \min_{y \in Y} \Bigl\|
E\big[p_\theta\big(x|E[q_\phi(z|\xtest) - y]\big)\big]\Bigr\|_2^2.\end{equation}

\subsubsection{Distance to the closest generated sample}
\label{opt}
A VAE is a generative model. Trained on a normal class only, one could expect that it will not be able to generate the datapoints that do not belong to the normal class. Thus a novelty score can be computed as a distance between an input and the closest sample that the VAE decoder is able to produce from any latent vector $z$:
\begin{equation}\label{opt1}N_{x-\hat{p}}(\xtest)=\min_{z}\left\|\xtest - E[p_\theta(x|z)]\right\|_2^2.\end{equation}
This is an optimization problem over the latent space. It can be solved using a non-linear optimization method such as L-BFGS~\cite{bfgs}. The encoder of the VAE with $\xtest$ as an input can be used to get an initial value for $z$ for optimization.

If the bottleneck of the VAE is not narrow enough, the VAE may still be able to reconstruct abnormal data. However, the values of $z$ for abnormal datapoints might be far from the unit Gaussian, since the VAE loss includes the term $\KL\big(q_\phi(z|x) \parallel \mathcal{N}(0,I)\big)$, and is trained on the normal data only. Thus the optimization can also be performed not over the whole latent space, but within some boundaries, defining a new novelty score metric $N_{x-\hat{p}_b}$. During the experiments, we found that the best results are achieved with $[-10, 10]$ boundaries for each of the dimensions of the latent space.

\subsubsection{Novelty as additive inverse of highest generated likelihood}
Another possibility is to not use Euclidean distance between the test sample and deterministic encoder output, but to compute the density value of \(p_\theta(\xtest|z)\). The novelty score in this case is equal to
\begin{equation}\label{opt2}N_{\hat{p}}(\xtest)=\min_{z}-{\log p_\theta(\xtest|z)}.\end{equation}
In comparison to the $N_{\bar{p}(\hat{q})}$ metric (Eq.~\eqref{esod}), where \(z\) is sampled from the approximated posterior \(q_\phi(z|\xtest)\), here \(z\) can take any value, or be restricted to lie within some boundaries for the same reasons as for the $N_{x-\hat{p}}$ score, Eq.~\eqref{opt1}.

\subsection{Novelty as full VAE loss}
If the VAE is trained on the ``normal'' class only, its loss function (Eq.~\eqref{vae_loss}) can be considered as a novelty metric itself. From a machine learning perspective, novel data that were not available during training will cause a high value of the model's loss function. From a probabilistic perspective, the VAE loss function value from datapoint $x_i$ is a reverse of the lower bound of the marginal probability $p(x_i)$ of this datapoint~\cite{vae1,vae2}. Novel datapoints have a low probability according to the model trained on a ``normal'' class and thus high loss value. Thus we propose a novelty metric:
\begin{equation}
N_{-\mathrm{ELBO}}(\xtest) = L_\mathrm{VAE} = -E_{q_\phi(z|\xtest)}[\log p_\theta(\xtest|z)] + \KL\big(q_\phi(z|\xtest)\parallel \mathcal{N}(0,I)\big).
\end{equation}
Like in the original VAE training algorithm~\cite{vae1}, the Monte Carlo estimate of $E_{q_\phi(z|\xtest)}$ should be computed. We also propose to use several samples from the probabilistic decoder $q_\phi(z|\xtest)$ and define another novelty metric:
\begin{equation}
N_{-{\widehat{\mathrm{ELBO}}}}(\xtest) = \min_{z\sim q_\phi(z|\xtest)}[\log p_\theta(\xtest|z)] + \KL\big(q_\phi(z|\xtest)\parallel \mathcal{N}(0,I)\big).
\end{equation}

\section{Experiments}
\label{results}

\begin{table}
  \caption{AUC scores of multiple sclerosis lesion segmentation for different q-space novelty detection methods. Many of our methods are at par with or outperform (marked in \textbf{bold}) the original method~$N_{x-y}$.}
  \label{results-table}
  \centering
  \begin{tabular}{lllll}
    \toprule
    \cmidrule(r){1-5}
    & & \multicolumn{3}{c}{AUC scores} \\
    Method & dim $z$ & scan 1 & scan 2  & scan 3 \\ 
    \midrule
    Baseline ($N_{x-y}$)& -- & 0.859 & 0.838 & 0.884 \\
    $N_\textrm{VAE-reg}$ & 16 & 0.803 & 0.773 & 0.812 \\
    $N_{\bar{q}-\bar{q}_y}$ & 16 & \textbf{0.893} & 0.831 & \textbf{0.890} \\
    $N_{q-q_y}$ & 16 & \textbf{0.888} & 0.835 & \textbf{0.893} \\
    $N_{q_Y}$ & 24 & \textbf{0.860} & 0.815 & 0.880 \\
    $N_{x-\bar{p}(\bar{q})}$ & 24 & 0.857 & \textbf{0.841} & 0.882 \\
    $N_{\bar{p}(\bar{q})}$ & 16 & 0.843 & \textbf{0.860} & 0.865  \\
    $N_{x-\bar{p}(\hat{q})}$ & 24 & 0.857 & 0.826 & 0.860 \\
    $N_{\bar{p}(\hat{q})}$ & 16 & 0.850 & \textbf{0.860} & 0.865 \\
    $N_{x-\hat{p}(\bar{q})}$ & 24 & 0.851 & \textbf{0.846} & 0.881 \\
    $N_{x-\hat{p}(\hat{q})}$ & 16 & 0.854 & 0.826 & 0.859 \\
    $N_{\bar{p}(\bar{q})-\bar{p}(\bar{q}_y)}$ & 24 & \textbf{0.895} & 0.829 & \textbf{0.891}\\
    $N_{\bar{p}(\bar{q})-y}$ & 16 & \textbf{0.893} & \textbf{0.852} & \textbf{0.897}\\
    $N_{p(\bar{q}_Y)}$ & 16 & 0.753 & 0.587 & 0.649 \\
    $N_{x-\bar{p}(\bar{q}_y)}$ & 16 & 0.830 & 0.811 & 0.857 \\
    $N_{p(\bar{q}_Y)(x)}$ & 16 & 0.760 & 0.600 & 0.652 \\
    $N_{x-\hat{p}}$ & 8 & 0.857 & \textbf{0.849} & 0.878 \\
    $N_{x-\hat{p}_b}$ & 8 & \textbf{0.859} & \textbf{0.849} & 0.879  \\
    $N_{-\mathrm{ELBO}}$ & 16 & \textbf{0.895} & \textbf{0.856} & 0.852 \\
    $N_{-{\widehat{\mathrm{ELBO}}}}$ & 16 & \textbf{0.897} & \textbf{0.867} & 0.873 \\
    \bottomrule
  \end{tabular}
\end{table}

In this section we evaluate the proposed novelty detection methods on multiple sclerosis lesion detection  in diffusion MRI and on the MNIST dataset. For both experiments we used an autoencoder with a bottleneck. We split the normal dataset into a training and validation set and used early stopping to prevent overfitting. We used the area under the curve (AUC) of the receiver operating characteristic as the quality metric because it quantifies the sensitivity and specificity of novelty detection across all possible decision thresholds, is robust to class imbalance, and is the most common accuracy metric for novelty detection in literature.

\begin{figure}
  \centering
  \includegraphics[width=1.0\textwidth,trim={4cm 0.5cm 4.2cm 2.4cm},clip]{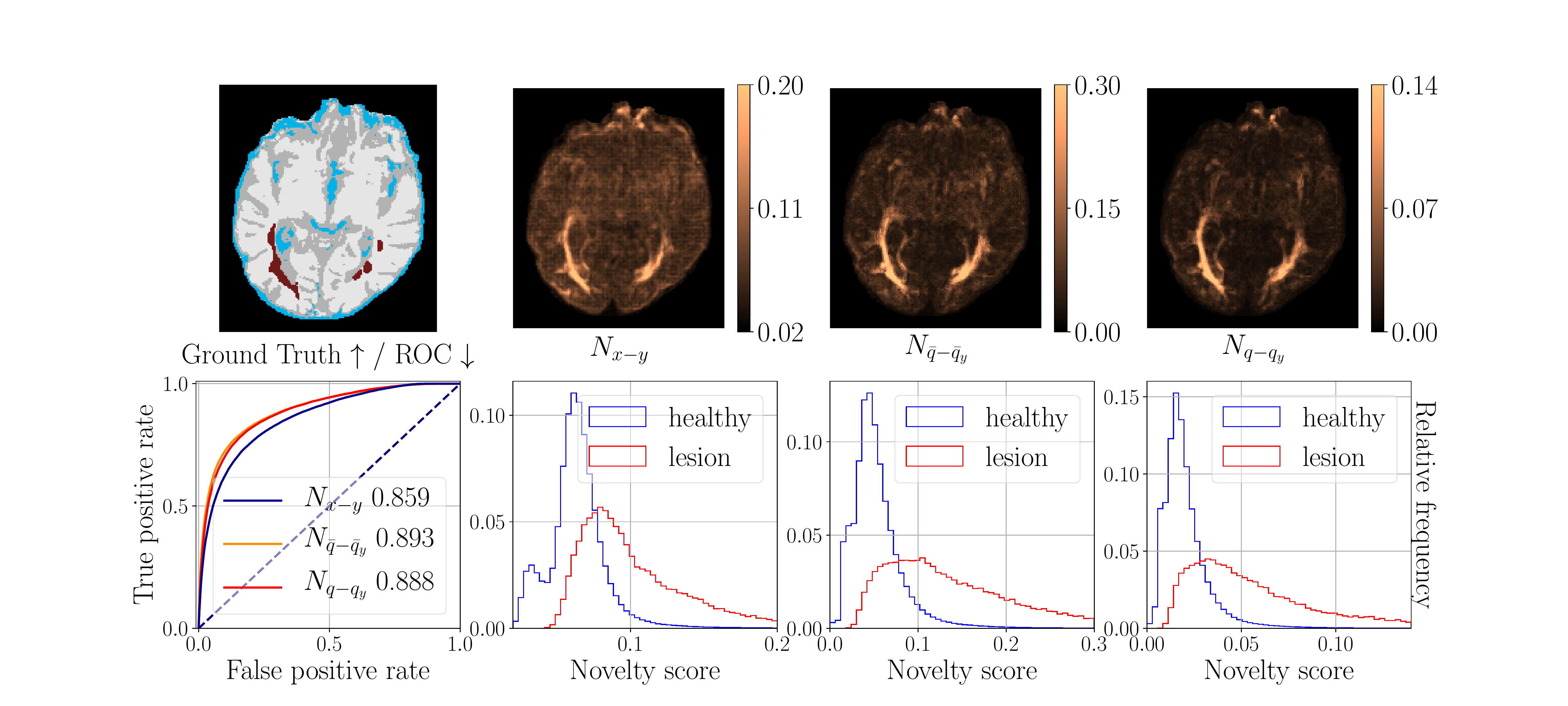}
  \caption{Feasibility of q-space novelty detection for multiple sclerosis lesion segmentation. Top row: manual lesion segmentation; novelty scores using the baseline method $N_{x-y}$ and two VAE latent space distance-based novelty detection methods $N_{\bar{q}-\bar{q}_y}$ and $N_{q-q_y}$. Bottom row: ROC for one scan measuring coincidence of novelty detection with human-marked labels (proposed methods outperform the baseline); normalized histogram of the novelty score for lesion and non-lesion voxels for the methods above (many lesion voxels have considerably higher novelty scores than healthy voxels). In other words, disease-related microstructural tissue changes are detected in a data-driven way, without any prior knowledge about them.
  }
  \label{segm}
\end{figure}

\subsection{Experimental setup for q-space novelty detection}
As the ``normal'' data to train the model for diffusion MRI, we used $26$ diffusion MRI scans of healthy volunteers that were split into 20 training and 6 validation scans. Each scan has six $b=0$ images and $40$ uniformly distributed diffusion directions ($b_\mathrm{max}=1200\mathrm{s/mm^2}$, SE-EPI, voxel size $1.8\mathrm{mm}\times1.8\mathrm{mm}\times2.4\mathrm{mm}$, matrix $128\times128$, $57$ slices, $T_E=94.5\mathrm{ms}$, $T_R=16\mathrm{s}$, distortion-corrected with elastix~\cite{elastix} and upsampled to $256\times256\times172$). Using machine learning nomenclature, we refer to these $46$ volumes (six $b=0$ and $40$ diffusion-weighted volumes) as \emph{channels} or \emph{voxel-wise features}. 
Test data consisted of three multiple sclerosis patients diffusion MRI scans with the same scan parameters as for healthy volunteers. Note that every voxel (rather than every scan) is a sample. Hence, the test set contains more than 5~million brain voxels in total, 55~thousand of which are lesion (abnormal) voxels. To validate the results of proposed methods we used multiple sclerosis lesion labels created by human raters using additional structural $T_2$-weighted scans. We compared the performance of proposed novelty detection algorithms with the distance-based q-space novelty detection algorithm~\cite{qnd, qnd2018} (described in Section~\ref{dmri}).

In order to avoid potential discrepancies in image intensity across scans, each scan was divided by its mean intensity. In addition, to prevent some channels from dominating over others, feature scaling was performed, i.e.~each channel was divided by the channel-wise mean taken across all scans. Each voxel in the diffusion MRI scan was considered as a separate data sample. Thus the local neighborhood information was ignored and only q-space measurements were used to perform ``voxel-to-voxel'' inference. Trained in this way, the VAE is able to produce the latent space and the original feature space distributions of the input data sample.

For each of the novelty metrics described in Section~\ref{methods}, we performed a hyperparameter search varying the dimensionality of the latent space, the depth of the architecture and the number of hidden layers to find the model that achieves the highest AUC score. Adam optimizer with learning rate 0.001 and batch size of 32768 voxels was used during training. Three different models with the following number of input/hidden/output features per layer were selected: $46-64-32-16-\mathbf{8}-16-32-64-46$; $46-128-64-32-\mathbf{16}-32-64-128-46$; $46-128-64-48-\mathbf{24}-48-64-128-46$. Here $46$ is the number of input/output features of the autoencoder. Results are shown in Table~\ref{results-table}; these three models can be distinguished by the ‘dim $z$’ column (the dimensionality of the latent space: $\mathbf{8}$, $\mathbf{16}$, or $\mathbf{24}$). Some of the produced multiple sclerosis lesions segmentations together with the ground truth and the segmentation produced by the baseline are shown in Figure~\ref{segm}.

\subsection{Experimental setup for the MNIST dataset}
We consider one of the MNIST handwritten digits as novel and train on the remaining ones. 80 \% of the normal data represents the training data. The test data consists of the remaining 20 \% of the normal data as well as all of the novel data. With 10 possible novelty classes/digits, this results in 10 different experiments.

We found that a relatively shallow network consisting of convolutional/upconvolutional and max-pooling layers performed best. The architecture is given in Table~\ref{tab:mnist-arch}. All convolutional layers have a filter size of $3\times 3$ and a stride of $1\times 1$. Inputs to the convolutional layers are zero-padded to preserve the image size. We used an Adam optimizer with a learning rate of $10^{-5}$ and a batch size of $64$.
For each possible novelty digit, we compare the performance of our proposed methods with linear PCA, PCA with a Gaussian kernel (kPCA)~\cite{kpca} and a VAE-based state-of-the-art approach~\cite{rprob}. The results of the best-performing methods are shown in Table~\ref{results-table2}.

\begin{table}
\caption{The architecture used for novelty detection on the MNIST dataset; dim $z$ refers to the dimensionality of the latent space.}
\label{tab:mnist-arch} 
\begin{tabularx}{\textwidth}{ll CCCC }
\toprule
Network Part & Layer Type & Number of Channels & Filter Size & Stride & Output Dimensions \\
\midrule
Input & - & - & - & - & $28\times 28$ \\\addlinespace
Encoder & Convolutional & 16 & $3\times 3$ & $1\times 1$ & $28\times 28$ \\
    & Maxpool & - & $2\times 2$ & $2\times 2$ & $14\times 14$  \\
    & Convolutional & 32 & $3\times 3$ & $1\times 1$ & $14\times 14$\\\addlinespace
Latent Space & Dense & - & - & - & dim $z$ \\\addlinespace
Decoder & Dense & - & - & - & $14\times 14$ \\
    & Convolutional & 32 & $3\times 3$ & $1\times 1$ & $14\times 14$ \\
    & Transp. Conv. & 16 & $2\times 2$ & $2\times 2$ & $28\times 28$  \\
    & Convolutional & 1 & $3\times 3$ & $1\times 1$ & $28\times 28$\\
\bottomrule
\end{tabularx}
\end{table}

\section{Discussion}
\label{discussion}

For the \textbf{diffusion MRI dataset}, most of the proposed methods show a good performance. More specifically, the method based on Euclidean distance in latent space ($N_{q-q_y}$ and $N_{\bar{q}-\bar{q}_y}$) outperforms the method based on Euclidean distance in the original data space ($N_{x-y}$; see also Figure~\ref{segm}), despite the fact that the model was not trained on abnormal data. This happens on one hand due to the fact that the VAE has a regularizer that keeps the latent representation of the normal data tightly clustered around the unit Gaussian, but at the same time, abnormal data that was not used during training can be mapped in a completely different region of the latent space (which is confirmed by the fact that $N_\textrm{VAE-reg}$ also produces meaningful novelty scores); and on the other hand due to the fact that the trained VAE simply happens to map abnormal data to slightly different latent-space regions. 
\begin{table}
  \caption{AUC scores for selected novelty detection methods on the MNIST dataset for all novelty classes/digits. Many of our methods outperform (marked in \textbf{bold}) the state of the art.}

  \label{results-table2}
  \centering
  \begin{tabular}{llllllllllll}
    \toprule
    \cmidrule(r){1-12}
    & & \multicolumn{10}{c}{AUC scores for novelty digits:} \\
    Method & dim $z$ & 0 & 1 & 2 & 3 & 4 & 5 & 6 & 7 & 8 & 9\\ 
    \midrule
    PCA~\cite{kpca} & -- & .785 & .205 & .798 & .632 & .682 & .627 & .733 & .512 & .493 & .410 \\
    kPCA~\cite{kpca} & -- & .694 & .231 & .801 & .638 & .702 & .598 & .720 & .560 & .502 & .420 \\
    An\&Cho~\cite{rprob} & -- & .917 & .136 & .921 & .781 & .808 & .862 & .848 & .596 & \textbf{.895} & .545 \\
    \cmidrule(r){1-12}
    $N_\textrm{VAE-reg}$ & 64 & .395 & .146 & .637 & .498 & .459 & .144 & .358 & .626 & .792 & .512 \\
    $N_{x-\bar{p}(\bar{q})}$ & 24 & \textbf{.941} & \textbf{.332} & .887 & .768 & .559 & .738 & .673 & .514 & .712 & .467 \\
    $N_{\bar{q}-\bar{q}_y}$ & 64 & \textbf{.922} & \textbf{.528} & \textbf{.965} & \textbf{.877} & \textbf{.870} & \textbf{.876} & \textbf{.872} & \textbf{.803} & .772 & \textbf{.637} \\
    $N_{q-q_y}$ & 64 & \textbf{.925} & \textbf{.480} & \textbf{.966} & \textbf{.882} & \textbf{.869} & .855 & \textbf{.868} & \textbf{.815} & .802 & \textbf{.641} \\
    $N_{\bar{p}(\bar{q})-\bar{p}(\bar{q}_y)}$ & 48 & \textbf{.935} & \textbf{.472} & \textbf{.958} & \textbf{.797} & .759 & \textbf{.869} & \textbf{.863} & \textbf{.744} & .805 & \textbf{.601} \\
    $N_{p(\bar{q})-p(\bar{q}_y)}$ & 48 & \textbf{.926} & \textbf{.501} & \textbf{.954} &  \textbf{.792} & .765 & \textbf{.866} & .847 & \textbf{.737} & .784 & \textbf{.597} \\
    \bottomrule
  \end{tabular}
\end{table}

Reconstruction-based methods also perform well, since the reconstruction of abnormal data (of which the model has not seen any during training) can be very imprecise. However, in order to be able to reconstruct the normal data, a VAE should have a sufficient number of latent dimensions to avoid significant information loss. The results of the methods $N_{x-\hat{p}}$ and $N_{x-\hat{p}_b}$ show the generative ability of the VAE. High AUC scores prove that the model is able to generate only the normal data it used during training. However, to achieve this ability, the bottleneck of the VAE (in contrast to reconstruction-based methods) should be narrow enough, otherwise the model may also be able to generate some ``random'' data that may be very close to the abnormal data in the original feature space. Finally, the VAE loss itself as the inverse of the lower bound of the likelihood of the data sample is a good novelty metric: abnormal data (none of which were used during training) will have low likelihood according to the model and thus high novelty score.

For the \textbf{MNIST dataset}, the methods showcase a greater variance in performance. Again, the distance-based methods applied to the latent space ($N_{\bar{q}-\bar{q}_y}$, $N_{q-q_y}$)  yield the best results, followed by the methods applied to the output-feature space ($N_{\bar{p}(\bar{q})-\bar{p}(\bar{q}_y)}$, $N_{p(\bar{q})-p(\bar{q}_y)}$). All of them clearly surpass the performance of existing approaches.
The VAE reconstructions of images are often blurry, which may explain the worse results of the reconstruction-based methods ( see $N_{x-\bar{p}(\bar{q})}$ in Table~\ref{results-table2}) for some novelty digits.
The VAE regularizer ($N_\textrm{VAE-reg}$) does not prove to be a sufficient novelty metric for most novelty digits.

\section{Conclusions}
\label{conclusion}

In this work, we presented a set of novelty detection algorithms that utilize the ability of variational autoencoders to reveal the internal structure of the normal class data. We found that high novelty scores produced by our methods coincide with multiple sclerosis lesions in diffusion MRI data, and many of our methods outperform the baseline \mbox{q-space} novelty detection method. Additionally, we evaluated the methods on the MNIST dataset, where we were able to surpass other state-of-the-art approaches. 

Interestingly, performance on novelty digit 1 was considerably worse than on the other digits both for previous methods and for out methods. As noted by \cite{rprob}, this likely is due to the fact that the more complex digits consist of similar, `1'-like structures that are implicitly learned by the VAE, in spite of it not encountering any novel data during training. We conclude that all methods are prone to failure if novel features have (partially) been learned from normal data.

\subsection*{Acknowledgments} 
This work was supported by the ERC Consolidator Grant ``3DReloaded''.

\end{document}